\documentclass[journal]{IEEEtran}

\usepackage{graphicx}
\usepackage{booktabs}
\usepackage{array}
\usepackage{cite}
\usepackage{xcolor}
\usepackage{amsmath}
\usepackage{amssymb}
\usepackage{setspace}
\usepackage{hyperref}
\begin{document}

\title{Agentic AI Networking for Heterogeneous Unmanned Aerial Systems in Low-Altitude Wireless Networks}

\author{Nguyen Duc Minh Quang,~\IEEEmembership{Graduate Student Member,~IEEE},
        Chang Liu,~\IEEEmembership{Member,~IEEE},\\
        Shuangyang Li,~\IEEEmembership{Member,~IEEE},
        and Derrick Wing Kwan Ng,~\IEEEmembership{Fellow,~IEEE}
\thanks{N.~D.~M.~Quang and C.~Liu are with the School of Computing, Engineering, and Mathematical Sciences, La Trobe University, Melbourne, VIC, Australia (e-mail: quang.nguyen@latrobe.edu.au; c.liu6@latrobe.edu.au).}%
\thanks{S.~Li is with the Chair of Communications and Information Theory, Technical University of Berlin, 10623 Berlin, Germany (e-mail: shuangyang.li@tu-berlin.de).}
\thanks{D.~W.~K.~Ng is with the School of Electrical Engineering and Telecommunications, University of New South Wales, Sydney, NSW, Australia (e-mail: w.k.ng@unsw.edu.au).}%
}

\maketitle

\begin{abstract}

Low-altitude wireless networks (LAWNs) are emerging as a key infrastructure for heterogeneous unmanned aerial systems that support concurrent services within a shared three-dimensional airspace. Their coexistence creates strong coupling among mobility, connectivity, and shared network resources, while heterogeneous services impose distinct and time-varying requirements. These interactions naturally form a dynamic non-cooperative game in which both operating conditions and coordination objectives evolve over time.
Conventional optimization and learning-based controllers typically rely on predefined objectives, limiting their ability to adapt autonomously to changing service requirements and resource priorities. To address this challenge, we propose a hierarchical hybrid large language model (LLM)– multi-agent reinforcement learning (MARL) architecture organized as a dual-loop structure. Specifically, an outer adaptation loop employs LLM-assisted game orchestration to interpret service requirements and operator intent, and reconfigure objectives and resource priorities, while an inner loop executes decentralized, parameter-conditioned MARL policies under the configured game. A logistics–monitoring case study illustrates how the proposed framework facilitates coordinated coexistence among heterogeneous services, adapting to evolving operating conditions without retraining the underlying MARL policies. Finally, we discuss key challenges and research directions toward scalable, trustworthy, and adaptive agentic LAWNs.

\end{abstract}

\begin{IEEEkeywords}
Agentic AI, swarms, low-altitude wireless networks, multi-agent
reinforcement learning, large language models, channel knowledge
map, game theory.
\end{IEEEkeywords}

\section{Introduction}
\label{sec:intro}

\begin{figure*}[t]
  \centering
  \includegraphics[width=0.7\linewidth]{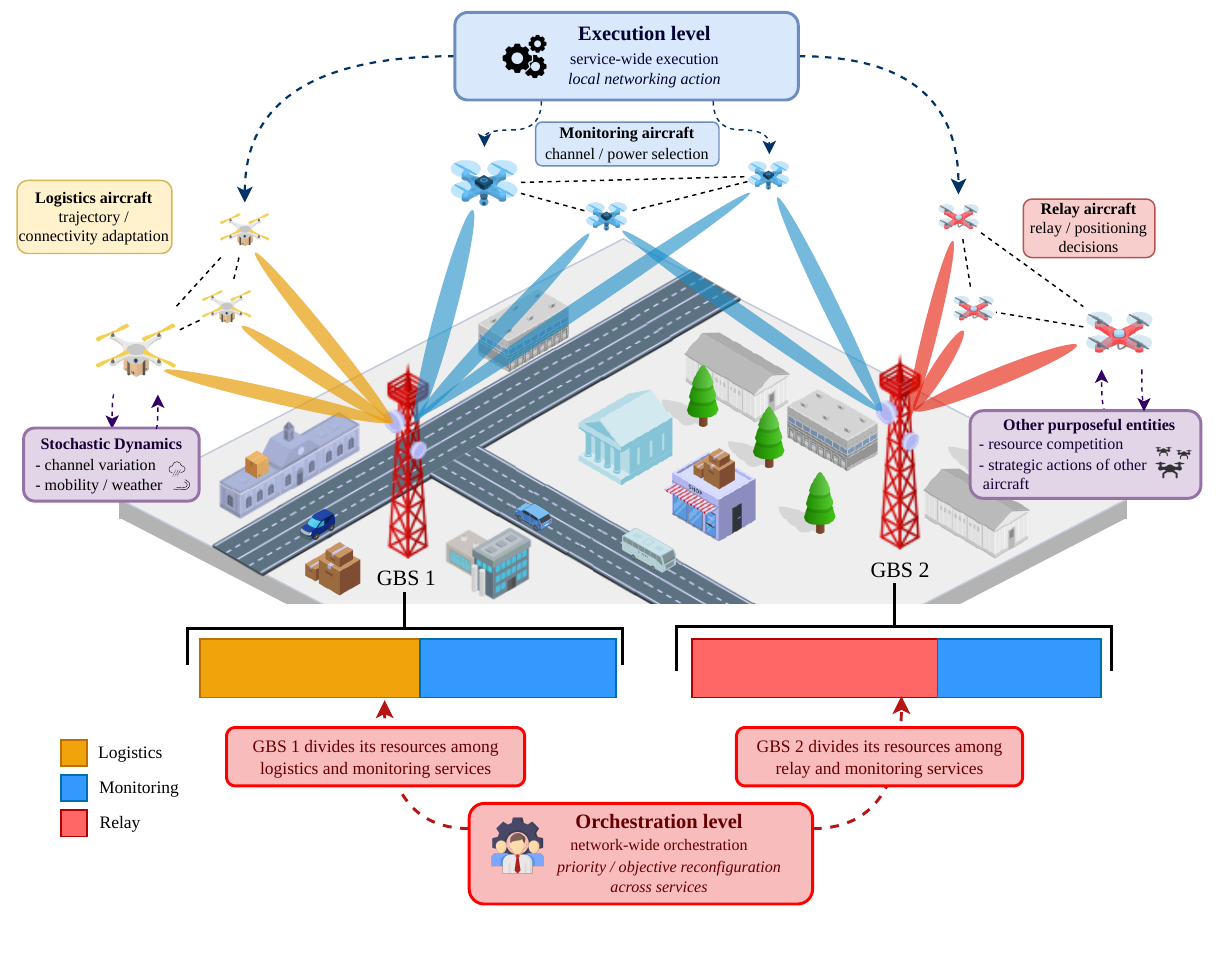}
  \caption{Heterogeneous services in the low-altitude economy contend for shared spectrum, ground-station capacity, and airspace. The bar beneath each GBS shows its resource divided among the services it currently serves.}
  \label{fig:vision}
\end{figure*}

The rapid expansion of low-altitude economic activities, from urban parcel delivery to continuous environmental monitoring, is driving the deployment of low-altitude wireless networks (LAWNs) that must maintain reliable connectivity for dense, highly mobile aerial systems operating below one kilometer~\cite{luo2026toward, jun2026low}. In fact, low-altitude aerial platforms are integral to the network topology, as their mobility dynamically affects link availability and association with ground base stations (GBSs), motivating learning-based designs that anticipate such variations to sustain reliable links~\cite{liu2026scalablepredictive}. Crucially, heterogeneous fleets (e.g., delivery, monitoring, relay) must share airspace and the finite communication resources of their serving GBS, i.e., the resources of bandwidth or transmit power each GBS allocates among the services it serves, as illustrated in Fig.~\ref{fig:vision}.

In practice, each service is typically engineered as a self-contained system with its own controller, coordination scheme, and communication stack~\cite{wu2026toward}. This isolation leaves cross-service conflicts unresolved, as each system optimizes its own objective without considering impacts on others, potentially causing congestion and mutual interference.
Since independently operated systems are coupled through shared resources, their interactions are naturally modeled as a game, with the objective of achieving stable and efficient coexistence. Importantly, the desired operating point of the game is dynamic, requiring reconfigurations of the runtime objective as the demands and priorities of the service shift. 
Concurrently, fast-changing topologies demand low-latency, decentralized control to avoid prohibitive coordination overhead. Meeting both needs motivates an autonomously adaptive framework that combines real-time distributed execution with high-level orchestration.

Learning-based wireless systems have developed perception~\cite{liu2019deep}, prediction~\cite{liu2022learning}, and adaptation~\cite{liu2020deeptransfer} as task-specific capabilities, but each serves as a foundational building block rather than autonomous closed-loop decision-making. Agentic AI is a natural candidate for the aforementioned requirements, since an agentic system is defined by an autonomous closed loop that adapts its own behavior based on observed outcomes. 
Specifically, two dominant types of agents naturally address these demands: large language model (LLM)-based agents interpret language objectives and contextual shifts, whereas multi-agent reinforcement learning (MARL) executes decentralized networking and mobility actions. This complementarity motivates a hybrid LLM–MARL framework that combines high-level reasoning with distributed policy execution.
However, existing agentic networking frameworks typically focus on single-service scenarios or fixed coordination objectives~\cite{xiao_toward_2025,zhao_agentification_2026}. Therefore, a critical gap remains in how to integrate these complementary agent capabilities so that heterogeneous services can autonomously adapt their coordination strategies as priorities and network conditions evolve.

To fill this gap, we propose a hierarchical hybrid LLM–MARL framework for adaptive coordination in heterogeneous LAWNs. The key contributions of this work are summarized as follows:

\begin{itemize}
    \item In contrast to isolated per-service optimization, we formulate execution-level coexistence in heterogeneous LAWNs as a dynamic non-cooperative game, and identify the orchestration-level requirement, providing a principled basis for coordinated coexistence over shared resources.

    \item To address the limitation of conventional fixed-objective control, we propose a hierarchical hybrid LLM–MARL architecture that couples LLM-assisted game orchestration with parameter-conditioned MARL execution, enabling runtime reconfiguration of coordination objectives without retraining the underlying decentralized policies, supported by a shared channel knowledge map.

    \item We present a logistics--monitoring coexistence case study to demonstrate how the proposed framework mitigates selfish resource competition and adapts the operating point to changing service priorities and network conditions, and discuss open challenges toward scalable agentic LAWNs.
\end{itemize}

\section{Heterogeneous-Service LAWNs}
\label{sec:agentic}

\subsection{Two Levels of Networking Tasks}
\label{sec:system}

Services in a LAWN are supported by swarms with different objectives. As illustrated in Fig.~\ref{fig:vision}, these swarms interact through shared spectrum, ground-station capacity, and airspace. Consequently, the environment faced by each aerial platform is shaped not only by stochastic dynamics (e.g., weather, channel variation) but also by the decisions of other purposeful entities~\cite{wu2026toward}. We organize the resulting networking tasks into two coupled operational levels: the \emph{execution level}, which supports service-wide networking, and the \emph{orchestration level}, which coordinates objectives across services. These levels differ in the nature of their outputs, the information they require, and their decision-latency requirements.

At the \emph{execution level}, aerial platforms autonomously select and execute networking actions to support their services under the currently configured objectives. These actions encompass three subtasks~\cite{wang2025toward}:
\begin{itemize}
    \item \emph{Spectrum and interference management} covers dynamic spectrum access, transmit-power control, and interference coordination.
    \item \emph{Topology and connectivity management} covers routing, relay selection, topology control, and handover across multiple GBSs and non-terrestrial platforms.
    \item \emph{Communication-aware mobility} covers trajectory planning that accounts for communication requirements.
\end{itemize}
Although organized around individual services, these actions remain coupled across swarms through shared resources. Aircraft can cruise at tens of meters per second through dynamic urban environments, causing link qualities and neighbor sets to change rapidly. These conditions impose stringent decision-latency requirements, including millisecond-scale responsiveness for time-critical networking actions. Execution relies primarily on on-board observations, including channel state, received power, and neighbor reports, and produces numerical control variables for communication and mobility.

At the \emph{orchestration level}, network-wide coordination involves allocating tasks and priorities among services, reconfiguring protocols, and coordinating with air-traffic management and human operators. Its inputs include service requirements, contracts, regulations, and operator intent, often expressed in natural language, together with feedback on network conditions and service performance. Orchestration operates on a slower timescale, typically seconds to minutes, revising service objectives, priorities, and protocol settings as demands and operating conditions change. Crucially, it does not prescribe individual networking actions; instead, it configures the objectives and operating settings under which aerial platforms select those actions.

The distinction is therefore functional as well as temporal: execution determines networking actions under the current objectives, whereas orchestration configures and revises those objectives across services. The two levels are coupled through downward guidance of execution and upward feedback on the resulting network conditions and service performance.

\subsection{Game Formulation}
\label{sec:game}

A strategic game provides a natural model for the execution-level interactions among aerial platforms~\cite{quang2026gametheoretic}. Unlike a centralized formulation with a single network-wide objective, it represents the distinct objectives and decision-making autonomy of different services. Swarms compete for limited shared resources, so their performance outcomes depend on one another's actions. This game-theoretic formulation captures strategic interdependence, supports analysis of network stability, and provides a structured basis for configuring coordination objectives across services. We characterize the game through four components~\cite{han2019game}:
\begin{itemize}
    \item \emph{Players:} Individual aircraft act as autonomous players. Aircraft supporting the same service share a service-specific utility weight and are modeled as cooperating toward a common service objective, whereas interactions across services are non-cooperative.
    
    \item \emph{Action space:} Aircraft select from common types of networking actions, including transmit-power control, relay selection, and communication-aware trajectory planning. These actions create spectral and spatial coupling across services through shared resources.
    
    \item \emph{Payoffs:} Each player's utility rewards service-specific performance, such as delivery speed or sensing data freshness, while penalizing excessive consumption of shared network resources. The resulting payoff depends on both the player's own actions and those of other aircraft.
    
    \item \emph{Information structure:} Each player selects its actions using local observations and may exchange information with neighboring aircraft. Players lack complete knowledge of network-wide resource states and the strategies and payoff functions of aircraft belonging to other swarms.
\end{itemize}

The orchestrator is not a player in this execution-level game: it neither selects the aircraft-level actions defined above nor has a player payoff within the game. Instead, it operates at the orchestration level to configure payoff parameters, such as service-specific utility weights, in accordance with Section~\ref{sec:system}. The information structure can be enriched by aggregating observations from multiple aircraft and sharing the resulting information with relevant players.

\subsection{Limits of Conventional Approaches}
\label{sec:limit}

While the game formulation captures strategic interactions among services, its practical realization requires both timely action selection and adaptive objective reconfiguration. Conventional solution approaches face different limitations at the execution and orchestration levels.
\begin{itemize}
    \item \emph{Execution level:} Model-based game solving and optimization struggle under unmodeled environments and tight latency budgets. While learning-based methods handle environment stochasticity and opponent dynamics, centralized execution incurs prohibitive computational and signaling overhead as fleet size and joint decision spaces grow~\cite{liu2023predictive}. The challenge is therefore to enable scalable, decentralized action selection under partial observability while accounting for cross-service interactions.

    \item \emph{Orchestration level:} The challenge is to revise the objectives governing execution, rather than merely optimize actions under fixed objectives. Operator intent, network conditions, and service performance are often expressed in natural language and must be interpreted to configure service objectives and payoff parameters. Because fixed rules and task-specific models scale and generalize poorly, adaptive orchestration requires context-aware interpretation beyond static mappings.
\end{itemize}

These complementary requirements motivate an agentic approach that closes two coupled loops: one adapting networking actions under the current objectives, the other adapting the objectives themselves.

\subsection{Agentic AI: Two Decision Mechanisms}
\label{sec:substrates}

\begin{table*}[t]
\caption{LLM-based agents versus MARL agents for agentic networking.}
\label{tab:substrates}
\centering
\small
\begin{tabular}{p{2.9cm}p{6.6cm}p{6.6cm}}
\toprule
\textbf{Dimension} & \textbf{LLM-based agents} & \textbf{MARL agents}\\
\midrule
Goal specification & Natural-language instructions & Reward function\\
\midrule
Knowledge & Broad pretrained knowledge & Narrow, acquired from the reward\\
\midrule
Adaptation & In-context, zero/few-shot learning & Policy update by trial and error\\
\midrule
Coordination & Natural-language negotiation & Reward shaping, peer communication\\
\midrule
Decision latency & $\sim$100~ms to seconds & Microseconds to milliseconds\\
\midrule
Explainability & States why it chose an action & Outputs an action only\\
\midrule
Deployment cost & High per-inference computational cost; large model footprint & High offline training cost; low on-board inference cost\\
\midrule
Best-served task level & Orchestration level: service and utility design, protocol reconfiguration, operator interface & Execution level: spectrum and interference, topology and connectivity, communication-aware mobility\\
\bottomrule
\end{tabular}
\end{table*}

To realize the closed-loop adaptation described above, an agentic system can combine specialized agents with complementary capabilities~\cite{sapkota2025ai}. Their suitability for different tasks depends on the computational model that primarily generates their decisions. Drawing on language-model-driven and reinforcement-learning-based agents~\cite{zhang_toward_2026}, we focus on two complementary decision mechanisms for heterogeneous-service LAWNs:
\begin{itemize}
    \item An \emph{LLM-based agent} uses a pretrained language model to interpret goals and contextual information, reason about possible responses, and generate structured decisions or tool calls. Natural-language instructions can be supplemented with observations and retrieved knowledge, while the context window and retrieval-augmented memory support reasoning across interactions~\cite{xu2024llmagent}. Its pretrained knowledge and contextual reasoning can support responses to unfamiliar instructions, although reliable generalization is not guaranteed. Decision generation, retrieval, and tool interactions also introduce computational and latency overhead.

    \item A \emph{MARL agent} selects actions using a policy learned through reward-driven interactions with other agents and the environment. Centralized training with decentralized execution is a common design, allowing each aircraft to act on local observations and available shared information~\cite{feriani2021marl}. Compact policy networks can support low-latency, on-board networking decisions. However, training can be data-intensive, and policy effectiveness depends on the reward specification and training experience. Changes beyond the trained objectives or operating conditions may therefore require further policy adaptation.
\end{itemize}
Table~\ref{tab:substrates} compares the two mechanisms across dimensions, including goal specification, adaptation, latency, and deployment cost. These characteristics are shaped by the underlying computational model, together with the training design, model size, and deployment platform.

These complementary capabilities motivate the hierarchical hybrid LLM--MARL architecture developed in Section~III. At the \emph{execution level}, MARL agents support service-wide networking through decentralized action selection under the current configuration. At the \emph{orchestration level}, LLM-based agents support network-wide coordination by interpreting evolving service requirements and performance feedback to reconfigure game parameters, such as service-specific utility weights.

\section{A Hierarchical Hybrid LLM–MARL Architecture}
\label{sec:architecture}

\subsection{Architecture Overview}
\label{sec:overview}

\begin{figure}[t]
  \centering
  \includegraphics[width=1\linewidth]{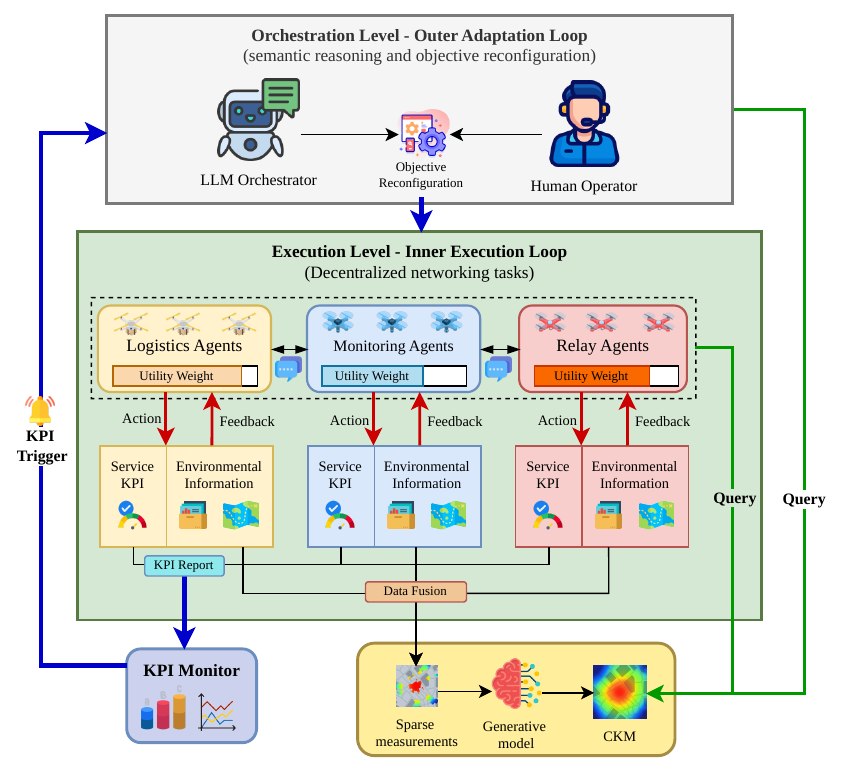}
  \caption{The closed-loop hierarchical LLM--MARL architecture.}
  \label{fig:architecture}
\end{figure}

The proposed architecture unites heterogeneous swarms through a closed perception--memory--reasoning--action (PMRA) loop, illustrated in Fig.~\ref{fig:architecture}. Building on the two-level decision structure identified in Section~II, it adopts a hierarchical hybrid LLM--MARL design organized as a dual-loop structure. The outer adaptation loop addresses orchestration-level coordination through LLM-assisted game orchestration, while the inner execution loop addresses execution-level networking through MARL-based decentralized mobility and communication decisions. The two loops are coupled through configurable game parameters and a shared channel knowledge map, allowing strategic adaptation and distributed execution to operate within a unified closed loop. This modular design also enables the shared memory (Section~III-B) and strategic reasoning (Section~III-C) components to operate either as standalone enhancements to existing controllers or jointly within the integrated PMRA framework.

In the inner execution loop, heterogeneous agents execute parameter-conditioned MARL policies under service-specific utility weights configured by the orchestrator. Each agent continuously observes its service key performance indicator (KPI) and local environmental information, queries the shared channel knowledge map for broader radio-environment knowledge, and maps the resulting context into mobility and networking actions. The resulting KPI reports are collected by the KPI monitor, while local environmental measurements are fused and used to update the channel knowledge map. This shared memory mitigates partial observability across heterogeneous swarms and provides a common environmental context for decentralized decision-making.
In the outer adaptation loop, the KPI monitor detects sustained performance degradation or other changes in operating conditions and triggers the LLM orchestrator. Together with direct operator input and the shared channel knowledge map, the orchestrator interprets evolving service requirements and reconfigures the coordination objectives, including utility weights and resource-pricing parameters. The updated game configuration is then broadcast to the inner-loop agents, whose parameter-conditioned MARL policies adapt their behavior without offline retraining. Since channel knowledge map construction and semantic reasoning are computationally demanding, these functions are hosted by the GBS infrastructure, while individual aerial agents retain lightweight autonomy for real-time execution.



\subsection{Memory: A Shared Channel-Knowledge Map}
\label{sec:memory}

Channel knowledge maps (CKMs), i.e., location-indexed representations of the radio environment such as path gain or blockage probability~\cite{zeng2024ckm}, serve as a shared environmental memory for the agents, enabling them to query expected channel conditions at unvisited locations and thus plan beyond their immediate sensing horizons~\cite{nguyen2026vit}. In LAWNs, however, no single aircraft can collect sufficient data to construct a global map, and the computational burden of CKM generation exceeds on-board capabilities.


CKM construction is therefore offloaded to the GBS infrastructure. Each GBS aggregates sparse measurements from its served aircraft and employs a generative model to reconstruct a dense local CKM without exhaustive sampling~\cite{nguyen2026diffusion, liu2020deepresidual}, while the backhaul federates these local estimates into a consistent, broader map. Aircraft upload local measurements and download the map patches they query, and relay cached patches over transient one-hop links to neighbors temporarily outside GBS coverage. The map is updated periodically or on demand when fresh measurements persistently deviate from prediction.

Both loops draw on this memory, as shown in Fig.~\ref{fig:architecture}. In the inner loop, agents query expected relay gains, interference patterns, or communication conditions along candidate trajectories prior to execution, and because the memory is \emph{shared} across heterogeneous swarms, a logistics aircraft can utilize blockage data collected minutes earlier by a monitoring aircraft, directly mitigating the partial-observability constraints of MARL. In the outer loop, the same shared view gives the LLM orchestrator visibility into spatial traffic density and spectrum bottlenecks, grounding its adjustment of congestion prices and multi-service payoffs in current physical conditions.

\subsection{Reasoning: Configuring and Solving the Game}
\label{sec:reasoning}

\begin{figure*}
    \centering
    \includegraphics[width=1\linewidth]{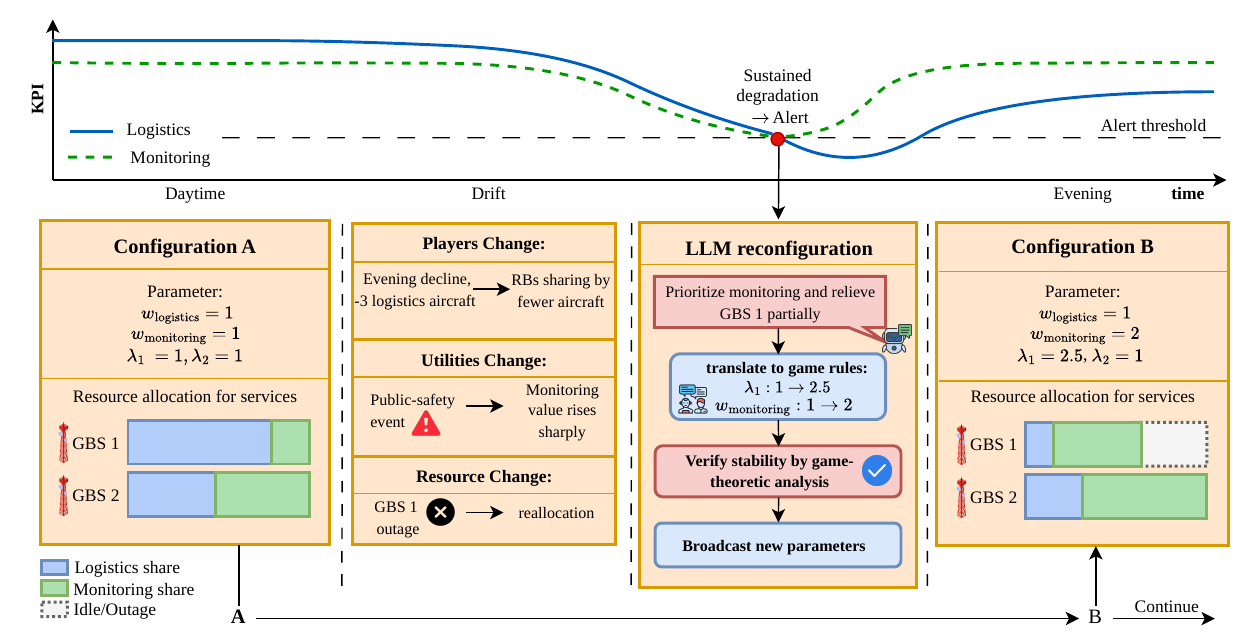}
    \caption{One reconfiguration cycle at the orchestration level.}
    \label{fig:game-theory}
\end{figure*}

\subsubsection{Game Configuration}
As established in Section~\ref{sec:game}, the interacting swarms form a game; the orchestration level's task is to configure the utilities that define it. Each service $m$ is credited for its primary key performance indicator ($\mathrm{KPI}$) while being charged for contested network resources~\cite{han2019game}, represented by a simplified payoff formulation:
\begin{equation}
  u_m = w_m\,\mathrm{KPI}_m - \sum_{r} \lambda_r\, c_{m,r},
  \label{eq:utility}
\end{equation}
where $\mathrm{KPI}_m$ denotes the service's performance metric (e.g., delivery rate or coverage freshness) pursued by aircraft of this service, $c_{m,r}$ is the consumption of shared resource $r$, and $\lambda_r$ and $w_m$ represent the resource price and utility weight, respectively. 
For example, Fig.~\ref{fig:game-theory} instantiates Eq.~\eqref{eq:utility} for service $m \in \{\mathrm{logistics}, \mathrm{monitoring}\}$ over the resources of two GBSs, $r \in \{1, 2\}$.
While real-world deployments may require complex models or hard constraints, this representative structure exposes three key configuration levers, including utility weights, congestion prices, and KPI definitions. These parameters serve as the primary operational variables adjusted by the orchestration level. Together, $(\boldsymbol{w}, \boldsymbol{\lambda})$ and the KPI definitions form the interface between the two loops: the outer loop selects them, and the inner loop executes under them.

\subsubsection{Inner Loop: Parameter-Conditioned MARL Execution}
\label{sec:marl-solver}
In the inner execution loop, each agent executes a decentralized MARL policy that maps local observations and CKM queries directly to real-time networking actions, such as waypoint selection, power control, and relay routing. The policies are trained via centralized training with decentralized execution, with action spaces tailored to specific tasks. To support runtime reconfiguration without continuous retraining, policies are trained over a parameterized distribution of game configurations. In particular, the serving GBS broadcasts the active parameter vector $(\boldsymbol{w}, \boldsymbol{\lambda})$ from Eq.~\eqref{eq:utility} as part of each agent's observation. This parameter-conditioned mechanism allows on-board MARL agents to adapt their behavior to the configuration issued by the outer loop, bypassing the need for offline retraining within the supported parameter set.

\subsubsection{Outer Loop: LLM-Assisted Game Orchestration}
\label{sec:orchestration}
What MARL cannot supply is the game itself, which is the task of the outer adaptation loop. In particular, the LLM translates operator intent, service contracts, and regulatory constraints, all expressed in natural language, into the weights, prices, and KPI definitions of Eq.~\eqref{eq:utility}. Reconfiguration is event-triggered, either by a KPI-monitor alert or by direct operator instruction, in response to drifts such as player-count change, shift in service value, or a change in resource availability. Before the parameters are broadcast, the candidate configuration is verified for stability, either analytically if the utility belongs to a structured game class (e.g., potential or congestion games) or via simulation and operator review. This preemptive check ensures stable convergence without relying on trial-and-error reward tuning. Fig.~\ref{fig:game-theory} illustrates one reconfiguration cycle, i.e., an evening logistics decline and a public-safety event pull the network off Configuration~A from daytime, and the monitor alerts once the degradation persists. The LLM then raises the monitoring weight $w_{\text{monitoring}} = 1 \xrightarrow{} 2$ and the congestion price $\lambda_1 = 1 \xrightarrow{} 2.5$ on the degraded GBS~1 while keeping $w_{\text{logistics}} = 1$ and $\lambda_2 = 1$, giving Configuration~B.

\section{Case Study: Logistics--monitoring Coexistence}
\label{sec:case}

\subsection{System Description}
\label{sec:case-setup}

To realize the proposed architecture, we consider an urban district served by multiple GBSs, over which two independently operated unmanned aerial vehicle (UAV) swarms cruise concurrently. A logistics swarm performs depot-to-customer deliveries, for which on-time completion depends on reliable control and coordination signaling from the serving GBS; a monitoring swarm moves slowly over designated zones, for which sustained coverage depends on continuous guidance to its assigned positions. The two are coupled through the GBS resources and airspace they share, yet neither service operator controls, or will accept control by, the other.

Both swarms are deployed under the proposed architecture of Section~\ref{sec:architecture}, with the competition for resource blocks (RBs) at the GBSs as the instantiation of Eq.~\eqref{eq:utility} adopted here. We compare against two reference baselines: a selfish operation setting, where each swarm optimizes its own utility without coordinating with the other; and a single controller with full information computes a centralized benchmark, which serves as an informative reference.

\begin{figure*}[th]
  \centering
  \includegraphics[width=1\linewidth]{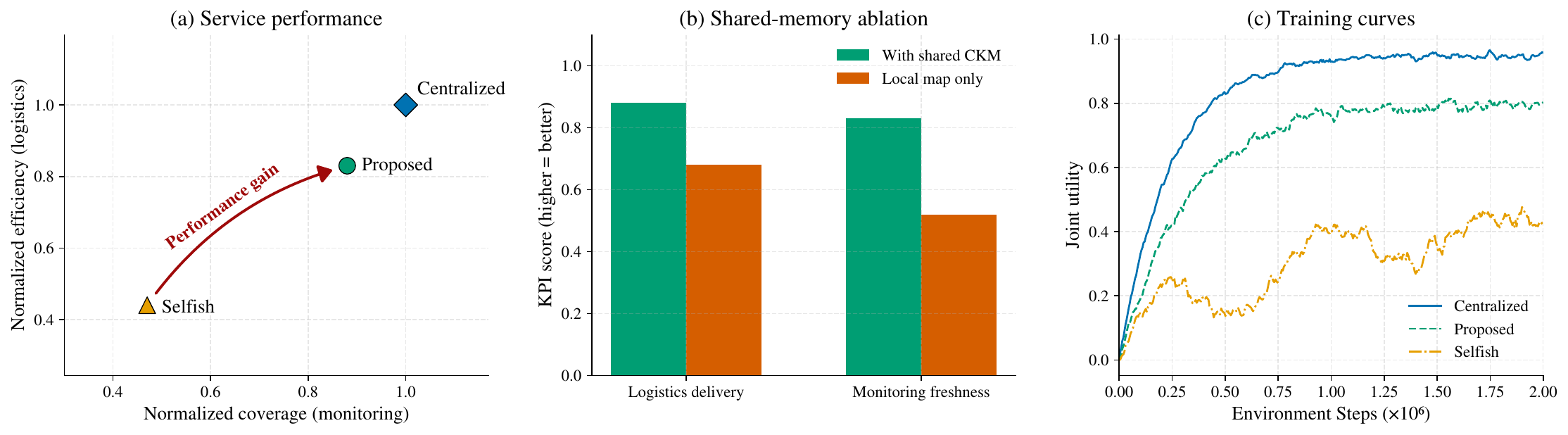}
  \caption{Illustrative logistics--monitoring coexistence results, evaluated across 5 independent random seeds and 200 test episodes per seed. (a) and (b) show testing-phase performance, while (c) shows the moving average of the joint utility during training. All scores are normalized by the centralized reference, so a value of 1.0 denotes parity and higher is better. The arrow in (a) marks the joint improvement of LLM-orchestrated MARL over selfish operation. In (b), the shared-memory ablation replaces the shared CKM with each agent's locally sensed map under a fixed game configuration.}
  \label{fig:results}
\end{figure*}

\subsection{Results and Discussion}
We simulate an urban district of roughly 1~km\textsuperscript{2} served by 3 GBSs, where 5 logistics UAVs and 5 monitoring UAVs operate concurrently. The airspace is discretized at a 10~m resolution up to an altitude of 300~m, UAVs move at up to 10~m/s, and each episode spans 100 time slots. Each GBS follows the 5G New Radio (NR) equal-RB allocation policy, dividing its RBs among the UAVs it currently serves. Each service's utility multiplies its average downlink rate by an efficiency factor, representing delivery efficiency for logistics and coverage persistence for monitoring. These efficiency factors instantiate $\mathrm{KPI}_m$ in Eq.~\eqref{eq:utility}, and $c_{m,r}$ is the number of RBs service $m$ draws from GBS $r$. Joint utility denotes $\sum_m u_m$ under the currently active configuration, normalized by the centralized reference so that a value of $1.0$ denotes parity. All scores are averaged over 5 seeds and 200 test episodes per seed.

Fig.~\ref{fig:results}(a) demonstrates that when the two swarms operate selfishly without coordination, the performance of both degrades. Specifically, they converge on the same favorably positioned GBS. Under equal-RB allocation, every UAV that joins a busy cell reduces the number of RBs of all connected UAVs, including its own. While individually rational, this selfish behavior results in a joint loss rather than a direct trade-off. By incorporating a congestion price, the orchestrator reflects the cost each service imposes on the others directly within the agent's utility function. Consequently, avoiding a crowded cell maximizes the agent's individual reward. Both services then attain roughly 85\% of the centralized reference without a centralized controller, corresponding to the joint improvement over selfish operation indicated in Fig.~\ref{fig:results}(a).

The contribution of the shared memory is evaluated separately. As shown in Fig.~\ref{fig:results}(b), restricting each UAV to its locally sensed map under a fixed game configuration reduces the normalized KPI by 0.20 and 0.30 for logistics and monitoring, respectively. Lacking a broader view of the radio environment, agents relocate based solely on local measurements and often fail to improve their positions. This penalty is larger for monitoring, which requires persistent coverage rather than a single discrete transfer. Fig.~\ref{fig:results}(c) illustrates these learning dynamics: the proposed policies converge smoothly, whereas the selfish baseline exhibits high fluctuation and settles at a significantly lower joint utility, as each swarm trains against unobservable competitors.

\subsection{Adaptation Under Runtime Reconfiguration}
\label{sec:ablation}

\begin{figure}
    \centering
    \includegraphics[width=1\linewidth]{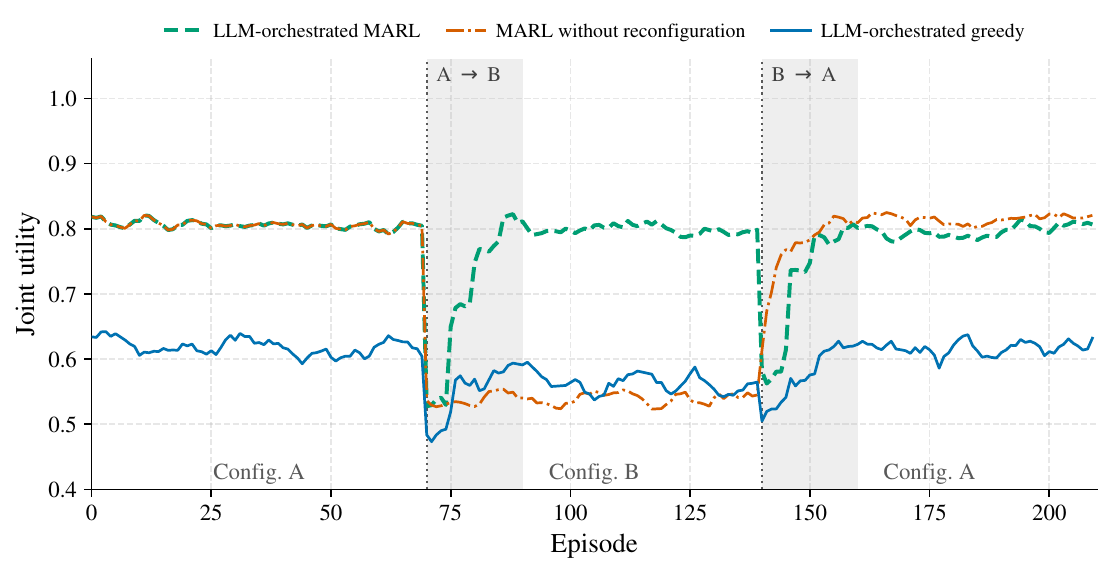}
    \caption{Closed-loop runtime reconfiguration. A public-safety event raises the resource price and the monitoring priority at episode~70; both revert at episode~140. Joint utility is evaluated under the currently active configuration and averaged over 5 seeds. Markers indicate the orchestration rounds.}
    \label{fig:adaptation}
\end{figure}

To verify the closed-loop adaptability of the proposed architecture, we undertake an ablation study. Under \emph{MARL without reconfiguration}, the game parameters are fixed at training time; under \emph{LLM-orchestrated greedy}, the orchestrator reconfigures as proposed, but the learned policies are replaced by a greedy strategy that maximizes the same configured utility at each slot~\cite{han2019game}. Configuration A sets $w_{\text{logistics}} = w_{\text{monitoring}} = 1$ and $\lambda_r = 1$ for all $r$. At episode 70, a simulated public-safety event leads the orchestrator to raise the monitoring weight to $w_{\text{monitoring}} = 2$ and the congestion price on the GBS to $\lambda_r = 1.5$ for all $r$, giving Configuration B; both revert at episode 140. Each orchestration round comprises one KPI report, one LLM inference, and one parameter broadcast, and the shaded bands in Fig.~\ref{fig:adaptation} mark the intervals over which the policies re-settle. All curves report joint utility under the currently active configuration.

As shown in Fig.~\ref{fig:adaptation}, the joint utility drops immediately at each event, since the active objective changes before any platform has relocated. The KPI monitor then triggers the LLM orchestrator, which infers a revised parameter vector from the reported KPIs and the event information, then refines it over successive rounds; recovery therefore proceeds in interpretable discrete steps and settles just below the pre-event level. Isolating the LLM orchestrator, the unreconfigured policies remain tuned to the original configuration and stay near $0.55$ throughout Configuration~B, recovering only at episode~140 when conditions revert to the one they were trained for. This suggests the LLM's capacity to interpret previously unseen events from their semantic content. The greedy baseline holds the refined objective but oscillates, settling approximately $0.20$ below the proposed architecture due to its per-slot decisions that ignore both future consequences and the reactions of other platforms. The two baselines thus fail differently, one holding the correct objective without reaching it and the other reaching a stale one precisely, so neither component suffices alone.

\section{Open Challenges and Future Directions}
\label{sec:challenges}

\emph{Scaling the equilibrium.} As the number and size of the swarm grow, training costs escalate, and stable operating points become harder to verify, compounding the LLM's difficulty in configuring congestion games. Deploying specialized LLM agents for distinct network segments or leveraging graph-based management can improve scalability by dividing the configuration workload. However, this distributed approach complicates human control and network verifiability, introducing the risk of cascading failures (e.g., LLM collapse) when autonomous orchestrators hallucinate or miscommunicate with neighboring segments.

\emph{The simulation-to-reality gap.} Policies trained in simulation often fail in real-world flight due to idealized platform dynamics (e.g., actuator delays, battery limits) and unmodeled environmental factors such as weather. Bridging this gap requires high-fidelity digital twins that co-model both the aerial platform and its operational environment. However, developing these comprehensive twins poses a significant challenge of its own, demanding intensive computational resources and extensive human engineering to accurately model the complex physical and aerodynamic edge cases required to mirror realistic environments.

\emph{Memory freshness and integrity.} A shared CKM represents a critical attack surface, as outdated or maliciously corrupted inputs can simultaneously misdirect all services. Proposed solutions include uncertainty-aware fusion and authenticated contribution mechanisms, ensuring agents prove their reliability before updating the shared map. However, the processing delays required for data encryption and identity verification directly conflict with the tight millisecond-latency requirements of the on-board wireless stack, creating a persistent trade-off between network security and real-time responsiveness.

\emph{Integrated terrestrial and non-terrestrial networks.} Low-altitude swarms will increasingly transition among terrestrial GBSs, high-altitude platforms, and satellite constellations. Consequently, satellites can provide a broad view that can stabilize and calibrate the detailed CKM assembled from lower-altitude platforms. Hierarchical federated learning offers a pathway to fuse these multi-scale representations, yet the severe propagation delays inherent to satellite links make real-time model synchronization highly challenging.

\emph{Interoperability across platforms.} The proposed architecture presupposes that aerial platforms, GBSs, and non-terrestrial platforms can exchange measurements, access the shared CKM, and coordinate their policies. At present, these systems are typically engineered by different vendors with incompatible interfaces, so there is neither a common memory to consult nor a unified strategic game to configure for a heterogeneous system. Emerging standards, such as those of the Global UTM Association (GUTMA) for traffic-management data exchange~\cite{GUTMA2026interop}, represent preliminary progress, but interface definitions capable of conveying rich environmental knowledge and enabling cross-service coordination are still lacking.

\section{Conclusion}
\label{sec:conclusion}

This article formulated heterogeneous LAWN networking as a non-cooperative game among aerial services, whose parameters are revised at runtime by an orchestration level. Conventional optimization and static learning-based methods prove inadequate for such dynamic interactions at scale. To address this limitation, we proposed a hierarchical hybrid LLM–MARL architecture, in which service utilities and resource prices are reconfigured through LLM-assisted reasoning over KPI feedback and operator intent, and realized through parameter-conditioned MARL policies grounded in a shared channel knowledge map, enabling coordination objectives to evolve without retraining the underlying policies. The logistics–monitoring case study demonstrated that competing services can attain stable, coordinated coexistence through this division of roles. Nevertheless, scaling to large swarms, closing the simulation-to-reality gap, and establishing standardized cross-service interfaces remain open challenges toward realizing fully autonomous agentic LAWNs.
\bibliographystyle{IEEEtran}
\bibliography{references}

\end{document}